\documentclass[letterpaper]{article} 
\usepackage{aaai25}  
\usepackage{times}  
\usepackage{helvet}  
\usepackage{courier}  
\usepackage[hyphens]{url}  
\usepackage{graphicx} 
\usepackage{natbib}  
\usepackage{caption} 
\usepackage{amsmath}
\usepackage{amssymb}
\usepackage{pifont}
\usepackage{algorithm}
\usepackage{algorithmic}

\usepackage{newfloat}
\usepackage{listings}
\DeclareCaptionStyle{ruled}{labelfont=normalfont,labelsep=colon,strut=off} 
\floatstyle{ruled}
\newfloat{listing}{tb}{lst}{}
\floatname{listing}{Listing}
\title{GoBERT: Gene Ontology Graph Informed BERT for Universal Gene Function Prediction}
\author{
    Yuwei Miao\textsuperscript{\rm 1},
    Yuzhi Guo\textsuperscript{\rm 1},
    Hehuan Ma\textsuperscript{\rm 1},
    Jingquan Yan\textsuperscript{\rm 1},
    Feng Jiang\textsuperscript{\rm 1},
    Rui Liao\textsuperscript{\rm 2},
    Junzhou Huang\textsuperscript{\rm 1}\thanks{Corresponding author: jzhuang@uta.edu}
}
\affiliations{
    \textsuperscript{\rm 1}Department of Computer Science and Engineering, University of Texas at Arlington\\
    \textsuperscript{\rm 2}Johnson \& Johnson Innovative Medicine\\

}

\usepackage{bibentry}

\begin{document}

\maketitle

\begin{abstract}
Exploring the functions of genes and gene products is crucial to a wide range of fields, including medical research, evolutionary biology, and environmental science. However, discovering new functions largely relies on expensive and exhaustive wet lab experiments. Existing methods of automatic function annotation or prediction mainly focus on protein function prediction with sequence, 3D-structures or protein family information. In this study, we propose to tackle the gene function prediction problem by exploring Gene Ontology graph and annotation with BERT (GoBERT) to decipher the underlying relationships among gene functions. Our proposed novel function prediction task utilizes existing functions as inputs and generalizes the function prediction to gene and gene products. Specifically, two pre-train tasks are designed to jointly train GoBERT to capture both explicit and implicit relations of functions. Neighborhood prediction is a self-supervised multi-label classification task that captures the explicit function relations. Specified masking and recovering task helps GoBERT in finding implicit patterns among functions. The pre-trained GoBERT possess the ability to predict novel functions for various gene and gene products based on known functional annotations. Extensive experiments, biological case studies, and ablation studies are conducted to demonstrate the superiority of our proposed GoBERT. 
\end{abstract}

%

\section{Introduction}
The accurate prediction of gene functions is a fundamental challenge in computational biology, with profound implications for medical research, evolutionary studies, and drug development. Understanding the roles of genes and their products is essential for understanding complex biological processes, which can lead to advancement in disease treatment and the development of novel therapeutics~\cite{jiang2024gte}. Gene Ontology (GO)~\cite{gene2019gene} and Gene Ontology Annotations (GOAs) are pivotal resources in this endeavor. GO is an organized graph containing the biological functions as the node and the relation between functions as the edge, providing a structured framework for classifying and annotating gene functions. 
For a certain gene or gene product, corresponding GOAs are selected nodes from GO, representing the functional roles. 
Despite their widespread use, these systems have not yet been fully explored for their potential in deep learning.

Ontologies are structured frameworks that define relationships between concepts within a domain, providing a standardized vocabulary for describing entities and their interconnections. In biological sciences, ontologies like GO~\cite{gene2019gene} help organize and analyze data by ensuring consistency across research studies and databases. GO categorizes gene functions into three domains: molecular function, cellular component, and biological process. The molecular function describes biochemical activities, the cellular component identifies locations within the cell, and the biological process outlines broader biological goals. GO is organized as a directed acyclic graph (DAG), where nodes represent specific gene functions (GO terms) in a coarse-to-fine manner, and edges denote relationships between these terms. Such structure allows for detailed descriptions of gene product roles, supporting the annotation process known as GOA. GO and GOA are essential for translating experimental results into structured notations, enabling enrichment analysis and gene function prediction. By integrating diverse datasets and facilitating cross-species comparisons, GO helps researchers explore the functional characteristics of genes, contributing to our understanding of biological processes in healthcare and disease diagnosis.


Traditionally, gene function prediction has primarily been based on proteins, with sequence data or three-dimensional structural information to infer the functional roles of proteins. While these methods have achieved success in certain contexts, they fall short of capturing the full spectrum of gene functions. Proteins, as products of gene expression, represent only a subset of the diverse functional roles that genes play in biological systems. Moreover, the reliance on protein-centric data limits the ability to generalize predictions across different species and functional categories. Common approaches include sequence-based methods that match the common sequence for functional annotation~\cite{altschul1990basic, altschul1997gapped}. Deep learning techniques such as convolutional neural networks, are employed to learn structural relationships and predict functions from sequence data~\cite{kulmanov2020deepgoplus}. Graph-based methods integrate sequence features with structural information to reveal relational patterns~\cite{jang2024accurate}. Multi-domain learning combines domain-specific and full-chain data to enhance prediction accuracy~\cite{zhou2022tasser, gligorijevic2021structure}. Despite these advancements, current methods often overlook implicit relationships between functions, highlighting the need for models that can uncover both explicit and implicit relations to provide a comprehensive understanding of gene functions.

To address these challenges, we introduce GoBERT (Figure~\ref{fig:framework}), an innovative BERT model that effectively captures both explicit and implicit relations within the GO DAG . For the explicit relations, we focus on two components: the semantic information of the raw text associated with each GO term and the structure information represented by the adjacency matrix of the whole GO DAG. We employ a self-supervised multi-label neighborhood prediction task using the adjacency matrix as labels and the Large Language Model (LLM) encoded text as initial embedding to capture these explicit relations. In addition to these easily observable explicit relations, GoBERT also targets implicit relations that may not be evident through semantic or structural information. These include phenomena such as pleiotropy, where a single gene can influence multiple phenotypic traits, leading to diverse functional outcomes across different biological processes. To achieve this, we design a self-supervised Masked Language Modeling (MLM) task that uncovers implicit relations by leveraging millions of genes with multiple annotations, enabling the prediction of novel functions through the designed masking strategy. This approach allows our GoBERT to bridge the gap between explicit and implicit functional relationships, further advancing our understanding of gene functions.  
The key contributions of this work are summarized as follows:
\begin{itemize}
    \item We introduce a novel research question and establish a benchmark for predicting novel functions of genes and gene products. To the best of our knowledge, this is the first study to apply deep learning methods for predicting novel gene functions solely based on known functions. 
    \item We propose the GoBERT model, which includes a self-supervised explicit relation pre-training task and an implicit relation pre-training task to comprehensively capture associations between the functions of genes and gene products. 
    \item We conduct experiments, case studies, and ablation studies to demonstrate the effectiveness of GoBERT in novel gene function prediction and its potential for biological applications. 
\end{itemize}

\section{Related Works}
\subsection{Automatic Function Annotation}
The evidence code of GOA reflects the annotation method of each function annotation, including six main categories: experimental, phylogenetically inferred, computational analysis, author statement, curator statement, and electronic annotation evidence code. Among these, only the GOAs with electronic annotation evidence code are automatic or semi-automatic annotated, all others require manual reviews or analysis which are time-consuming and costly. The InterPro2GO~\cite{burge2012manual} method automates protein function annotation by utilizing associations between GO terms and generalized sequence models of homologous protein groups, assigning GOA based on sequence matches.

Furthermore, many current automatic annotation methods often depend on additional information, which constrains their applicability within the protein function prediction. Traditional and widely used Basic Local Alignment Search Tool (BLAST)~\cite{altschul1990basic} rapidly identifies sequence similarities using heuristic algorithms. PSI-BLAST~\cite{altschul1997gapped} enhances sensitivity by iteratively refining searches with position-specific scoring matrices, exhibiting weak yet biologically relevant protein relationships. 
DeepGO~\cite{kulmanov2018deepgo} utilizes CNNs and Protein-Protein Interaction (PPI) networks to learn protein representations and predict protein functions hierarchically within Gene Ontology.
DeepText2GO~\cite{you2018deeptext2go} combines deep semantic text representation with sequence-based methods, integrating text-based classifiers to enhance large-scale protein function prediction across Gene Ontology domains.
DeepFRI~\cite{gligorijevic2021structure} is a graph convolutional network for predicting protein functions by integrating sequence features and structural data. 
I-TASSER-MTD~\cite{zhou2022tasser} uses deep learning for multi-domain protein structure and function prediction, enhancing assembly accuracy and annotation at both the domain and full-chain levels.
PhiGnet~\cite{jang2024accurate} utilizes statistics-informed graph networks to predict protein functions from sequences, employing evolutionary couplings and residue communities for functional site identification.
It is noteworthy that all these methods primarily focus on utilizing existing sequences or structural information to predict protein functions, overlooking the implicit relationships that may exist between the functions themselves.

\subsection{BERT-based Pre-train Models}
BERT~\cite{devlin2018bert} pre-training strategies are widely used across various fields. Originally, BERT as a text encoder learns the relationship between tokens and encodes input text as a context-aware embedding~\cite{liu2019roberta, lan2019albert}. BERT is capable of processing professional scientific text~\cite{beltagy2019scibert, alsentzer2019publicly}, and can be adapted to many domain-specific applications.
BioBERT~\cite{lee2020biobert} is a pre-trained BERT on large-scale biomedical corpora for biomedical text mining. 
SMILES-BERT~\cite{wang2019smiles} is pre-trained using a masked SMILES recovery task for molecular representation learning and is effective in downstream molecular property prediction. The scBERT~\cite{yang2022scbert} model utilizes a Performer-based architecture with a two-stage training process, involving self-supervised pre-training on unlabelled scRNA-seq data, followed by supervised fine-tuning on the cell type annotation datasets. DNABERT~\cite{ji2021dnabert} utilizes BERT architecture to model k-mer sequences in DNA, and DNABERT-2~\cite{zhou2024dnabert} enhances this approach by incorporating additional genomic features and structural information.
Our proposed GoBERT extends the strengths of BERT to the realm of gene and gene product function prediction by introducing and providing a novel method for interpreting complex biological datasets.

\section{Method}
\begin{figure*}[t]
\centering
\includegraphics[width=\textwidth]{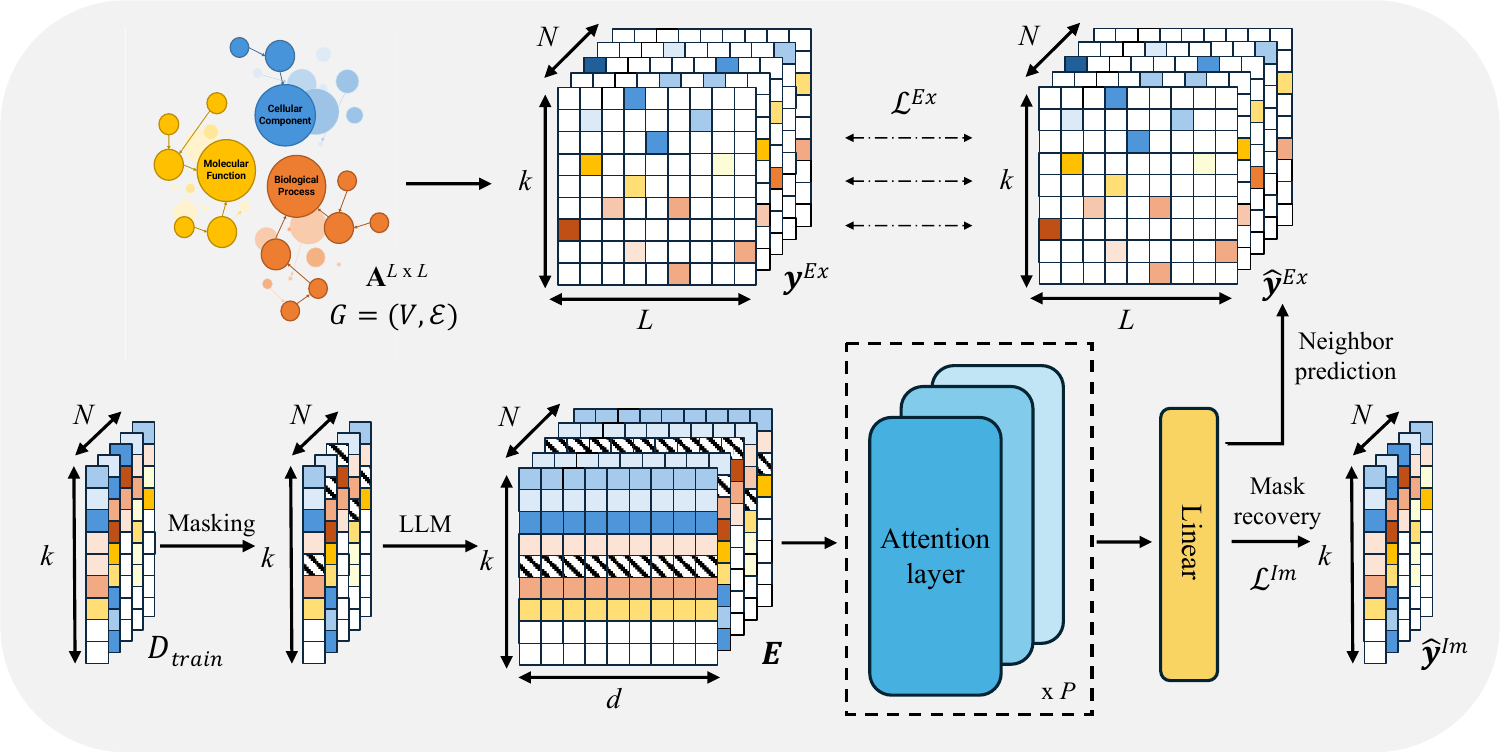} 
\caption{This figure illustrates the main components and framework of GoBERT. The white blocks represent pad tokens and the slashed blocks are mask tokens. Red, yellow, and blue blocks indicate functions belonging to Biological Processes, Molecular Functions, or Cellular Components categories, respectively. $D_{\text{train}}$ contains input data with N genes and k functions for each gene. Then, the designed masking strategy is applied. LLM-generated embedding is used in the initialization of token embedding $\mathbf{E}$ in GoBERT. For the implicit pre-train task, the $\mathcal{L}^{\text{Im}}$ is the loss between predicted mask functions and the ground truth functions in $D_{\text{train}}$. For the explicit pre-train task, labels $\mathbf{y}^{\text{Ex}}_i$ are obtained from adjacency matrix $\mathbf{A}$ of GO DAG, where $\mathcal{L}$ denotes the total number of nodes or functions. $\mathcal{L}^{\text{Ex}}$ is calculated for capturing the structural information of functions.}
\label{fig:framework}
\end{figure*}

In this section, we first formulate the problem, which aims to predict the gene and gene product functions using known functional annotations. We then introduce our proposed GoBERT in detail to tackle this problem, including the two specifically designed pre-training tasks: explicit and implicit relation prediction. The explicit relation task utilizes the semantic information of GO terms and the structure of GO DAG, while the implicit relation task captures the underlying biological relationships among functions using MLM. The overview of our framework is demonstrated in Figure \ref{fig:framework}. 

\begin{figure}[t]
\centering
\includegraphics[width=\columnwidth]{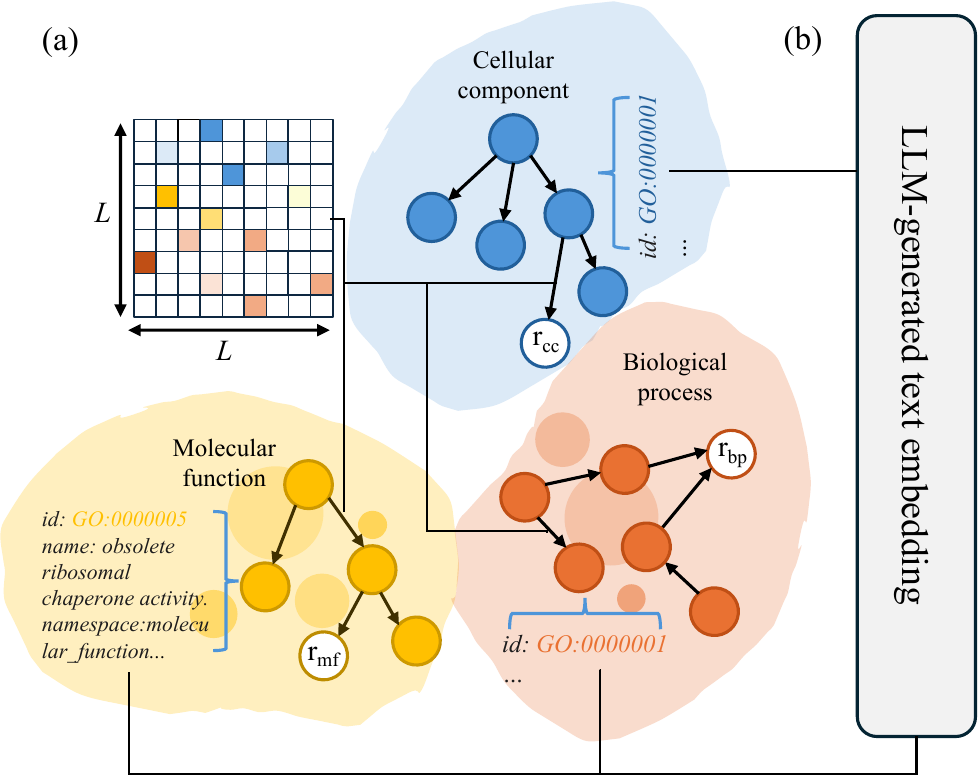} 
\caption{Explicit relations between functions are depicted through GO DAG structures (edges) and semantic information (nodes). (a) The structure information can be represented by the adjacency matrix of the GO DAG, which serves as labels in neighborhood prediction. (b) Semantic information is captured by encoding raw text descriptions of each node in GO DAG with LLMs.}
\label{fig:explicit}
\end{figure}

\begin{figure}[t]
\centering
\includegraphics[width=\columnwidth]{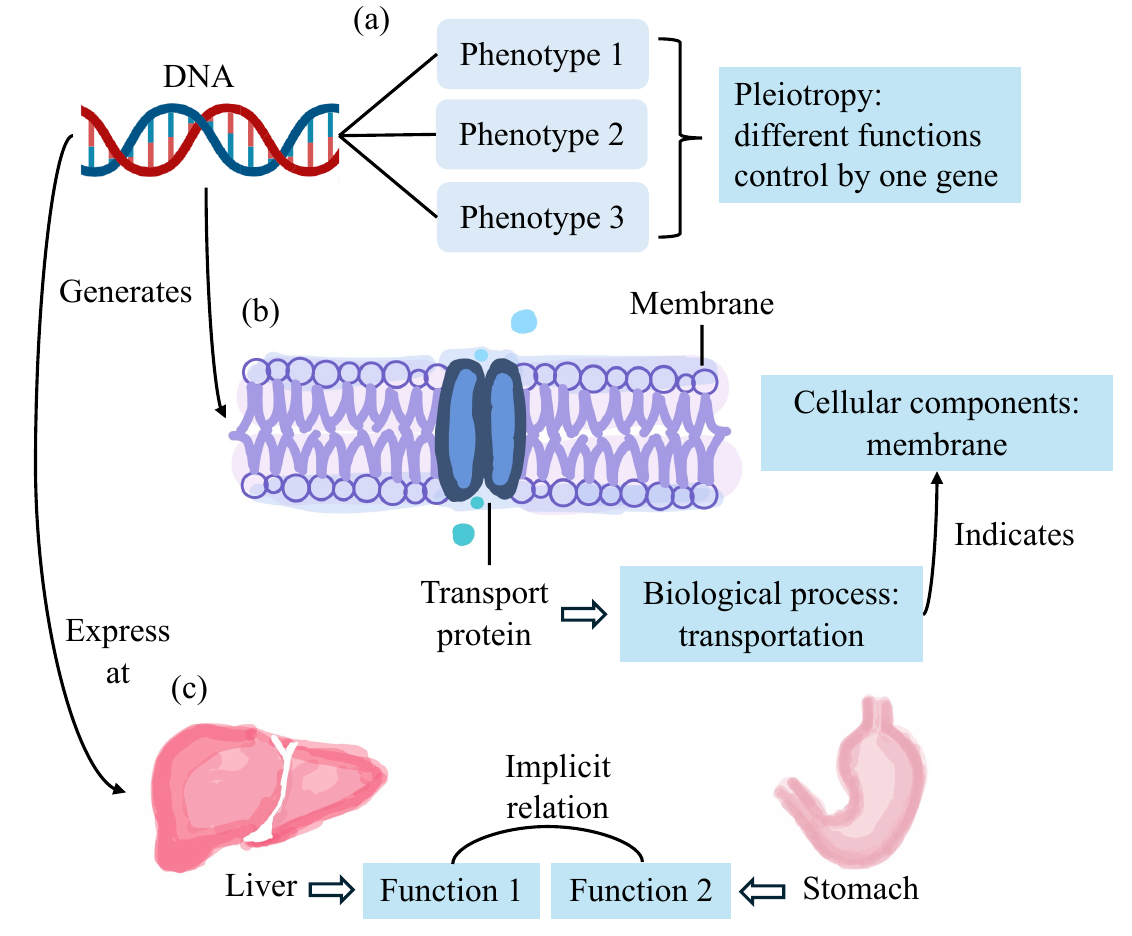}
\caption{Implicit relations among gene functions are demonstrated by three examples. (a) Pleiotropy: multiple phenotypes can be controlled by a single gene, indicating there are underlying relationships between these functions. (b) A protein that contributes to the transportation biological process is potentially located on the membrane. (c) The same gene can produce different functional outcomes depending on the expression tissue, gene products may result in different functions in the liver and stomach.}
\label{fig:implicit}
\end{figure}

\subsection{Problem Formulation}
We consider GO DAG as \( G = (V, \mathcal{E}) \) and its corresponding adjacency matrix as $\mathbf{A}^{L\times L}$, where \( V \) is the set of nodes representing GO terms. $\mathcal{E} \subseteq V \times V$ is the set of directed edges representing relationships between GO terms and $L = |V|$. For a gene denoted as $g_i$, where $i \in [1,N]$, the training data is defined as $D_{\text{train}} = \{V_i, T_i\}^N_{i=1}$ where $V_i$ is a subset of $V$ which contains $k$ GO annotations for gene $g_i$ and $T_i$ is a set of text describing each node in $V_i$. Set $V_i$ and $T_i$ can be formally defined as: 
\begin{align}
    V_i = \{ v_i^1, v_i^2, \dots, v_i^k\}, v_i^k \in V \quad T_i = \{t_i^1, t_i^2, \dots, t_i^k\}
\end{align}
where $t_i^k$ is the descriptive text that concatenates attributes such as id, name and namespace for node $v_i^k$: 
\begin{align}
    t_i^k = \operatorname{concat}(v_i^k.\text{id}, v_i^k.\text{name}, v_i^k.\text{namespace}, \dots)
\end{align}
and text $t_i^k$ is organized in a key-value pair format (see node text in Figure \ref{fig:explicit}). Given partial annotations of a gene $g_i$, the objective of our task is to train a model that is capable of predicting other gene annotations.

\subsection{Explicit Relation}
As illustrated in Figure~\ref{fig:explicit}, we consider the explicit relation of functions from two key components: the semantic information consists in the functional descriptions \( T_i \), and the GO DAG structure captured by the adjacency matrix $\mathbf{A}$. We introduce a self-supervised multi-label neighborhood prediction task for BERT with the adjacency matrix $\mathbf{A}$ as labels and the LLM-generated node embedding as the BERT token initial embedding.

\subsubsection{GO DAG structure}  
GO graph is a DAG with three main sub-graphs structuring different categories of gene functions. 
The acyclic nature of the DAG graph ensures that:
\begin{equation}
(v_i, v_j) \in \mathcal{E} \implies (v_j, v_i) \not\in \mathcal{E}.
\end{equation}

There are three root nodes for the three sub-graphs, $R = \{ r_{mf}, r_{bp}, r_{cc} \} \subseteq V$. Each GO term \( v_i \) belongs to exactly one of these roots. Formally, this can be expressed as:
\begin{equation}
\forall \, v \in V, \; \exists ! \, r \in \{r_{mf}, r_{bp}, r_{cc}\} : \operatorname{path}(v, r),
\end{equation}
where \(\operatorname{path}(v_i, v_j)\) indicates the existence of a directed path from node $v_i$ to node $v_j$.

Every edge \( (v_i, v_j) \in \mathcal{E} \) is characterized by one of the five types of relationships: \textit{is\_a}, \textit{part\_of}, \textit{regulates}, \textit{positively\_regulates}, and \textit{negatively\_regulates}. For example, ``lactase activity" is\_a ``hydrolase activity", ``regulation of DNA recombination" regulates ``DNA recombination”, and ``mitotic interphase” part\_of ``mitotic G1 phase”.

\subsubsection{Semantic Information} 
Text encoders that average the pre-trained word embedding~\cite{arora2017simple, li2020sentence} are proven to be effective for capturing semantic information. Text embedding generated by these approaches may be limited to the input text, while LLMs offer the potential to incorporate richer contextual information~\cite{wang2023improving}. Potentially, LLM-generated embedding can lead to a more informative encoding of gene function descriptions and enhance the ability to detect function relations from text descriptions.

The text descriptions $t_i^k$ for gene $g_i$ are encoded into text embedding vector $\mathbf{h}_i^k$ by LLM denoted as $\psi(\cdot)$.
\begin{equation}
(\mathbf{H}_i)^{d\times k} = [\mathbf{h}_i^1, \mathbf{h}_i^2, \dots, \mathbf{h}_i^k ], \; \mathbf{h}_i^k = \psi(t_i^k) \in \mathbb{R}^d
\end{equation}

\subsubsection{Neighborhood Prediction}
Inspired by GIANT~\cite{chien2022node}, GoBERT learns the explicit relations among functions by performing self-supervised neighborhood prediction, which formulate as a multi-label classification problem. For gene $g_i$, there are a set of node \( V_i \) with function text descriptions set \( T_i \) as gene annotations.

In GoBERT, the full vocabulary contains \( V \) and special tokens. 
The token embedding for gene $g_i$ is denoted with 
\begin{align}
    (\mathbf{E}_i)^{d\times k} = [\mathbf{e}_i^1, \mathbf{e}_i^2, \dots, \mathbf{e}_i^k], \; \mathbf{e}_i^k\in \mathbb{R}^d
\end{align}
where $d$ is the hidden size. The LLM-generated embedding $\mathbf{H}_i$ are used for token embedding initialization:
\begin{align}
    \mathbf{E}_i: {\mathbf{E}_i|} _{t=0} = \mathbf{H}_i
\end{align}

The GO DAG structure is represented by the adjacency matrix \( \mathbf{A} \in \{0, 1\}^{L \times L} \), where \( \mathbf{A}_{ij} = 1 \) indicates an edge between \( v_i \) and \( v_j \). The objective is to predict the neighborhood set \( \mathcal{N}(v_i) \) for a given node \( v_i \) by learning a function $f(\cdot)$ that outputs a probability distribution over the possible labels:
\begin{equation}
\mathbf{y}^{\text{Ex}}_i = \mathbf{A}_{i,:} \in [0, 1]^L \quad \hat{\mathbf{y}}^{\text{Ex}}_{i} = f(\mathbf{e}_i),
\end{equation}

where $\mathbf{y}_i$ is the ground truth label vector of node \( v_i \) and \( y_{ij} = 1 \) indicates the $j_{\text{th}}$ node is a neighbor of \( v_i \). The prediction function $f(\cdot)$ is trained by minimizing the following:
\begin{equation}
\mathcal{L}^{\text{Ex}} = \frac{1}{N} \sum_{i}^N\operatorname{BCELoss}\left (\mathbf{y}^{\text{Ex}}_i, \hat{\mathbf{y}}^{\text{Ex}}_i \right ),
\label{eq:explicit}
\end{equation}
where $\operatorname{BCELoss}(\cdot)$ is the Binary Cross-Entropy loss function and \(\hat{\mathbf{y}}^{\text{Ex}}_i\) is the predicted label vector. The prediction function \( f(\cdot) \) incorporates both semantic feature \( \mathbf{x}_i \) and structural context from \( \mathbf{A} \).

\subsection{Implicit Relations}
Besides the easy-to-observe explicit relations, there are implicit relations between the functions, as shown in Figure \ref{fig:implicit}. These implicit relations may not be semantically similar or included in the hierarchical structure represented by GO DAG. For examples in Figure~\ref{fig:implicit}, pleiotropy is the phenomenon where a single gene influences multiple phenotypic traits. This means that mutations or variations in a single gene can affect different, seemingly unrelated characteristics or biological processes. The same gene expressed in different tissues may also produce different functional results. One gene could participate in multiple biological processes and therefore regulate multiple pathways to achieve various functions. These implicit patterns among the functions of gene and gene products are difficult to observe and largely rely on wet lab experiments to discover. In this work, with millions of genes with multiple annotations, we design the self-supervised MLM task to unravel the implicit relations with the injected knowledge of explicit relations. 

\subsubsection{Masking Strategy}
In MLM, a certain percentage of tokens are typically masked at random for recovery. Due to the special structure of GO DAG, the original masking strategy have limitations on dataset $D_{\text{train}}$, which excessively simplify the recovery task. We designed a masking strategy for the datasets based on the GO DAG to overcome the challenges.

First, root nodes \(r_{mf}, r_{bp}, r_{cc}\) are excluded for masking. As these root nodes are universally valid functional annotations for all genes and gene products, masking and recovering the root nodes does not provide the desired information. Second, nodes with any predecessor in \( V_i \) are invalid for masking. 
In GO DAG, the relation ``A part\_of B" forms an edge from A to B, indicating A is the predecessor of B, if A exists in \( V_i \), it is easy to infer its masked successor node B. Masking and recovering B neither contributes to model learning nor provides an adequate evaluation during inference, as it may oversimplify the prediction task. Overall, the designed masking strategy is formulated as:
\begin{align}
    &V_i^{\text{mask}} = \{v_i^k \mid v_i^k \in V_i \backslash R, \; \operatorname{predec}(v_i^k) \cap V_i\ = \varnothing\},  
\end{align}
where \(\operatorname{predec}(v_i^k)\) denotes the set of predecessor nodes for \( v_i^k \).
The final mask is randomly sampled from \(V_i^{\text{mask}}\) with a masking rate of  $\alpha_{\text{mask}}$. When applying the mask, 80\% of the selected tokens are replaced with a mask token, 10\% with a random token, and the remaining 10\% are left unchanged.

\subsubsection{Novel Function Prediction}
To capture implicit relations between functions, we first remove the positional encoding, making an input order-invariant GoBERT. The goal is to train a BERT model $g(\cdot)$ that accurately predict the masked functions based on the context by minimizing the loss function \(\mathcal{L}_\text{Im}\), defined as:
\begin{equation}
\hat{\mathbf{y}}^{\text{Im}}_{i} = g(\mathbf{E}_i), \; {\mathbf{E}_i|} _{t=0} = \mathbf{H}_i,
\end{equation}
\begin{equation}
\mathcal{L}^{\text{Im}} = \frac{1}{N} \sum_{i}^N \operatorname{CELoss} (\mathbf{y}^{\text{Im}}_{i}, \hat{\mathbf{y}}^{\text{Im}}_{i}),
\label{eq:implicit}
\end{equation}
where \(\mathcal{L}_\text{Im}\) is a cross-entropy loss for classification, $\mathbf{H}_i$ represents the LLM-generated features from the unmasked context functions. 

\begin{table*}[htbp]
\centering
\caption{Novel Function Prediction Results: average and standard deviation of 5 runs with different masking seeds are reported.}
\label{tab:novel_function_prediction}
\begin{tabular}{lcccc}
\hline
\textbf{Method} & \textbf{Top-1 Acc} & \textbf{Top-5 Acc} & \textbf{Top-1 Acc w/ depth} & \textbf{Top-5 Acc w/ depth} \\
\hline

GoBERT w/o Neighborhood Prediction & 30.76 $\pm$ 0.43 & 53.91 $\pm$ 0.28 & 50.34 $\pm$ 0.40 & 73.53 $\pm$ 0.14  \\
GoBERT w/o Semantic Information & 32.31 $\pm$ 0.48 & 54.96 $\pm$ 0.14 & 51.83 $\pm$ 0.21  & 74.73 $\pm$ 0.13 \\
GoBERT w/o Masking Strategy & 28.03 $\pm$ 0.49 & 50.37 $\pm$ 0.30 & 47.67 $\pm$ 0.33 & 70.41 $\pm$ 0.06 \\
\hline
GoBERT & \textbf{34.08 $\pm$ 0.76} & \textbf{57.47 $\pm$ 0.53} & \textbf{53.91 $\pm$ 0.37} & \textbf{76.15 $\pm$ 0.24}  \\
\hline
\end{tabular}
\end{table*}

\section{Model Training}

\subsection{BERT}
BERT~\cite{devlin2018bert} is a multi-layer bidirectional Transformer that processes tokens using self-attention mechanisms to learn contextual relations. Usually, BERT-based model are pre-trained with masked language modeling (MLM) or next sentence prediction (NSP) task. In this study, we incorporate MLM for novel function prediction.  

\subsubsection{Scaled Dot-Product Attention}
Scaled dot-product attention calculates attention weights by scaling the query-key dot products and applying softmax for focus.
\begin{equation}
\text{Attention}(Q, K, V) = \operatorname{softmax}\left(\frac{QK^T}{\sqrt{d_k}}\right) V,
\end{equation}
where \(V\) is value, \(Q\) is query, \(K\) is key, and \(d_k\) is the dimension of \(K\).
\subsubsection{Multi-Head Attention}

Multi-head attention in BERT enables the model to focus on different parts of the input sequence simultaneously. It is computed as:
\begin{equation}
\operatorname{MultiHead}(Q, K, V) = \operatorname{Concat}(\text{head}_1, \dots, \text{head}_h)W^O,
\end{equation}
where each attention head is defined as:
\begin{equation}
\text{head}_i = \operatorname{Attention}(QW_i^Q, KW_i^K, VW_i^V).
\end{equation}
Here, \(W_i^Q\), \(W_i^K\), \(W_i^V\), and \(W^O\) are learnable parameter matrices.

\subsubsection{Position-wise Feed-Forward Networks}
The position-wise feed-forward networks apply two linear transformations with a ReLU activation:
\begin{equation}
\text{FFN}(x) = \max(0, xW_1 + b_1)W_2 + b_2
\end{equation}
\subsubsection{Layer Normalization}
Layer normalization is used to stabilize and accelerate training by normalizing inputs:
\begin{equation}
\operatorname{LayerNorm}(x) = \gamma \left(\frac{x - \mu}{\sigma}\right) + \beta,
\end{equation}
where \(\mu\) and \(\sigma\) are the mean and standard deviation of the input, and \(\gamma\) and \(\beta\) are learnable parameters.

\subsection{Training Objective}
GoBERT jointly optimizes the explicit and implicit tasks. The explicit task focuses on semantic and structural relationships in GO DAG and the implicit task uncovers hidden patterns among functions.
The final training objective combines both tasks with a balancing factor \(\lambda\) controlling their contributions:
\begin{equation}
\mathcal{L}^{\text{Total}} = \lambda \mathcal{L}^{\text{Ex}} + (1 - \lambda) \mathcal{L}^{\text{Im}}.
\end{equation}
Here, \(\lambda\) is a hyperparameter that balances the importance of the explicit and implicit objectives. \(\mathcal{L}^{\text{Ex}}\) is the objective for the explicit pre-train task described in Equation~\eqref{eq:explicit}, and \(\mathcal{L}^{\text{Im}}\) is the objective for the implicit pre-train task described in Equation~\eqref{eq:implicit}.

\section{Experiments}
\subsection{Experimental Setups}
\subsubsection{Dataset}
For the explicit neighborhood prediction task, we generate the dataset label based on the adjacency matrix of GO DAG with a down-sampling rate of 0.001 for balancing. We construct the implicit annotation prediction dataset from 5.1 million genes with 139 million functional annotations from UniEntrezDB~\cite{miao2024unientrezdblargescalegeneontology}. After filtering out duplicate gene annotation inputs, 413k genes with 6.1 million annotations are selected. We perform a K-Means partition on the gene embedding generated by averaged function embedding from the LLM to split this dataset as the train, valid, and test sets. In particular, the test set is used to generate experiments of novel function prediction. Dataset details are available in the Appendix.

\subsubsection{Evaluation Metrics}
Since there is more than one valid function prediction for each mask, we utilize the top-1 and top-5 accuracy with corresponding results in designated depth as evaluation metrics for the novel function prediction task. Top-k accuracy measures correctness by considering a prediction correct if the ground truth result is shown in the top k highest probability predictions. Depth is defined as the shortest path from a node to its root. Smaller depth indicates a more general/coarse function, while larger depth indicates a more specific/fine function. For each gene, there are valid function predictions at each depth, therefore we select the top-k function that has the highest k probabilities at the target depth.

\section{Experimental Results}
In this section, we perform comprehensive experiments to demonstrate the effectiveness of GoBERT. We pre-train 20 epochs for each model for the following experiments. First, GoBERT can predict novel functions for gene and gene products with only a few functional annotations. Second, we provided the case studies of TOF2 and MGT1 to illustrate how GoBERT novel function prediction is meaningful to biological studies. Finally, ablation studies are conducted to demonstrate the importance of each component in GoBERT. 

\subsection{Novel Function Prediction}
There are over 47k classes of functions in GO DAG and numerous genes across species in the natural world.
Conducting wet lab experiments for the functions of one gene is exhaustive, let alone experimenting with all functional classes of every gene. Our proposed GoBERT preserves the ability to conduct large-scale novel function predictions by utilizing known functions. GoBERT possesses the ability to generalize and predict novel functions for any gene or gene product since it is not restricted by the availability of specific additional information such as gene sequences or protein three-dimensional structures.  

Table~\ref{tab:novel_function_prediction} presents the main results of our GoBERT model. Notably, GoBERT achieves a 76.15\% accuracy with top-5 accuracy at specific depth. Given that multiple valid functions could correspond to the masked position, this high accuracy suggests that GoBERT effectively captures the relationships among functions through both explicit and implicit pre-training tasks. These promising results highlight the potential of deep learning to successfully address the novel function prediction task we introduced, which is critical for advancing biological studies.

\subsection{Case Studies}
In the case studies, we demonstrate the effectiveness of GoBERT in predicting gene functions that align with biological research. These capabilities highlight its potential to aid researchers in understanding complex biological systems and identifying novel functions beyond the reach of traditional methods. Detailed information on the gene and gene products used in the case studies is available in the Appendix. 

\subsubsection{Case 1: Inferring Cellular Components of TOF2 by Molecular Function and Biological Process}

Inferring cellular components is crucial for biological studies. While changes in phenotypes are more apparent and easier to observe, accurately tracking the actions or expression sites of gene products is essential for targeted treatments, especially in disease contexts. 
Our proposed GoBERT can assist in predicting the locations where these products act or are expressed within the cell, as demonstrated in the following case study.

TOpoisomerase I-interacting Factor (TOF2), a protein-coding gene with Entrez Gene ID 853880,  controls the required protein for rDNA silencing, mitotic rDNA condensation, and Cdc14p activation. It promotes rDNA segregation and condensin recruitment to the replication fork barrier. The input functions of TOF2 \( V_i \) can be break down as: 
\begin{itemize}
    \item Molecular Function: 
    \begin{itemize}
        \item GO:0005515 (protein binding, d=2)
        \item GO:0019211 (phosphatase activator activity, d=4)
        \item GO:0000182 (rDNA binding, d=7)
    \end{itemize}
    \item Cellular Component: 
    \begin{itemize}
        \item GO:0005739 (mitochondrion, d=3)
        \item GO:0005634 (nucleus, d=5)
        \item GO:0005730 (nucleolus, d=5)
    \end{itemize}
    \item Biological Process:
    \begin{itemize}
        \item GO:0000183 (rDNA heterochromatin formation, d=5)
        \item GO:0070550 (rDNA chromatin condensation, d=5)
        \item GO:0007000 (nucleolus organization, d=6)
        \item GO:0031030 (negative regulation of septation initiation signaling, d=6)
        \item GO:0034503 (protein localization to nucleolar rDNA repeats, d=7)
    \end{itemize}
\end{itemize}
where d is the shortest path from function nodes to the roots in the GO DAG, and a larger d indicates a more specific function. To show the pre-trained GoBERT possesses the ability to infer cross-category functions, we take the molecular functions and cellular components of TOF2 as input to predict its Cellular Component. In the results, GO:0005739 is the most probable function among 5,570 choices for d=3. GO:0005634 and GO:0005730 rank second and third, respectively, among 13,469 choices for functions in d=5. 

Moreover, GO:0005694 (chromosome), which ranks first place of the predicted cellular component at d=5, is also a reasonable novel function inferred by GoBERT. The resources of manual TOF2 function annotation cover a broad range of TOF2-related studies. However, the connection between TOF2 and the chromosome is identified through a study on factors associated with Chromosome Transmission Fidelity (CTF4), which might have been overlooked in the manual annotation process~\cite{villa2016ctf4}. This study demonstrates that the association of TOF2 with CTF4 is crucial for maintaining the size of chromosome 12, which contains rDNA repeat arrays.

\subsubsection{Case 2: Predict Predecessor Function}
As the functions in GO DAG vary from general to specific, which are denoted as nodes from root to predecessor. It is worth predicting which specific function a gene may have given the general function resulting from our experiments. 

The methylated-DNA--protein-cysteine methyltransferase (MGT1), a gene with Entrez ID 651327, involves in protecting against DNA alkylation damage and localizing to the peroxisome in a Pex5p-dependent manner~\cite{maglott2005entrez, xiao1991primary, sassanfar1990identification, david2022pls1}. The functions of MGT1 possess the following patterns:
\begin{itemize}
    \item Molecular Function: 
    \begin{itemize}
        \item GO:0003908 (methylated-DNA-[protein]-cysteine S-methyltransferase activity)
        \item GO:0003677 (DNA binding)
        \item GO:0005515 (protein binding)
    \end{itemize}
    \item Cellular Component:
    \begin{itemize}
        \item GO:0005634 (nucleus)
        \item GO:0005777 (peroxisome)
    \end{itemize}
    \item Biological Process:
    \begin{itemize}
        \item GO:0006281 (DNA repair) \textit{is\_a} GO:0006259 (DNA metabolic process)
        \item GO:0032259 (methylation)
    \end{itemize} 
\end{itemize}
Specifically, GO:0006259 are more general functions in GO DAG compared to its predecessor GO:0006281. We demonstrate that GoBERT can accurately infer more specific functions by using general functions as input to predict their predecessors. GO:0006281 (DNA repair) is the most probable predecessor of GO:0006259 (DNA metabolic process) among 39 predecessor choices. 

\subsection{Ablation Study}
In this section, we conduct the ablation studies of the main components of GoBERT on the novel function prediction task, including removing the explicit pre-training, removing the LLM-captured semantic similarity, and without the designed masking strategy. The results demonstrate the importance of each component in GoBERT. Table~\ref{tab:novel_function_prediction} shows the comparison of the main results of the ablation studies.

As observed, the masking strategy is crucial for maintaining the difficulty of the MLM task, enabling the model to generalize better in predicting novel functions. Without this strategy, the generalized ability of GoBERT decreases significantly. Semantic information, derived from LLM-encoded text, provides essential functional insights that help the model detect complex patterns. The neighborhood prediction task is also vital for GoBERT to be graph-informed. It allows GoBERT to utilize the structure relations among functions in GO DAG, which are fundamental for accurate function prediction. Overall, the ablation studies demonstrate that these components are crucial to GoBERT. Detailed ablation studies on parameter selection can be found in the Appendix.

\section{Conclusion and Future Work}
In this paper, we introduce GoBERT, a BERT model informed by gene ontology graphs, designed for universal gene function prediction. Explicit and implicit pre-train tasks capture informative relationships among the functions. We propose a new approach for investigating novel gene functions that rely on known functions and the implicit relationships between them, which provides the fundament for future research in gene function prediction studies. We conduct experiments, biological case studies, and ablation studies to demonstrate the performance and effectiveness of GoBERT. 

As we are the first to investigate novel gene function prediction sorely based on known functions, many research directions and challenges are still open and need further exploration. For example, the relationship between a function and a gene can be categorized as a gene is either confirmed to have a function, not to have a function, or a function is not yet experimented. Currently, we only consider the confirmed functions in model training. However, for generating a complete annotation set for a gene, incorporating functions that a gene is proven not to have is essential for future work. Furthermore, the GoBERT generates informative functional representations that can be integrated into various biological applications in the future. Introducing other modalities of data such as gene sequence and expression level could also further unravel the relationship between a certain gene and its functions. 

\section{Acknowledgements}
This work was partially supported by US National Science Foundation IIS2412195, CCF-2400785 and the Cancer Prevention and Research Institute of
Texas (CPRIT) award (RP230363).

\bibliography{aaai25}

\end{document}